\documentclass[lettersize,journal]{IEEEtran}

\usepackage{times}
\usepackage{epsfig}
\usepackage{graphicx}
\usepackage{amsmath}
\usepackage{amssymb}
\usepackage{gensymb}
\usepackage[table,xcdraw]{xcolor}
\usepackage{multirow}
\usepackage{enumitem}
\usepackage{caption} 

\usepackage{hyperref}
\hypersetup{
  colorlinks,%
  breaklinks=true,
  citecolor=blue,%
  filecolor=blue,%
  linkcolor=blue,%
  urlcolor=blue
}

\begin{document}

\title{\LARGE \bf
FisheyeDetNet: 360\degree~Surround view Fisheye Camera based Object Detection System for Autonomous Driving
}

\author{Ganesh Sistu$^{1\dag}$ and Senthil Yogamani$^{2\dag}$\\
$^1$University of Limerick, Ireland\\
$^2$Valeo Vision Systems, Ireland \quad $^\dag$co-first authors
}

\maketitle
\thispagestyle{empty}
\pagestyle{empty}
\begin{abstract}
Object detection is a mature problem in autonomous driving with pedestrian detection being one of the first deployed algorithms. It has been comprehensively studied in the literature. However, object detection is relatively less explored for fisheye cameras used for surround-view near field sensing. The standard bounding box representation fails in fisheye cameras due to heavy radial distortion, particularly in the periphery. To mitigate this, we explore extending the standard object detection output representation of bounding box. We design rotated bounding boxes, ellipse, generic polygon as polar arc/angle representations and define an instance segmentation mIOU metric to analyze these representations. The proposed model FisheyeDetNet with polygon outperforms others and achieves a mAP score of 49.5\% on Valeo fisheye surround-view dataset for automated driving applications. This dataset has 60K images captured from 4 surround-view cameras across Europe, North America and Asia. To the best of our knowledge, this is the first detailed study on object detection on fisheye cameras for autonomous driving scenarios.
\end{abstract}
\section{Introduction} \label{Intro}
\begin{figure*}
\centering
    \includegraphics[width=0.99\linewidth]{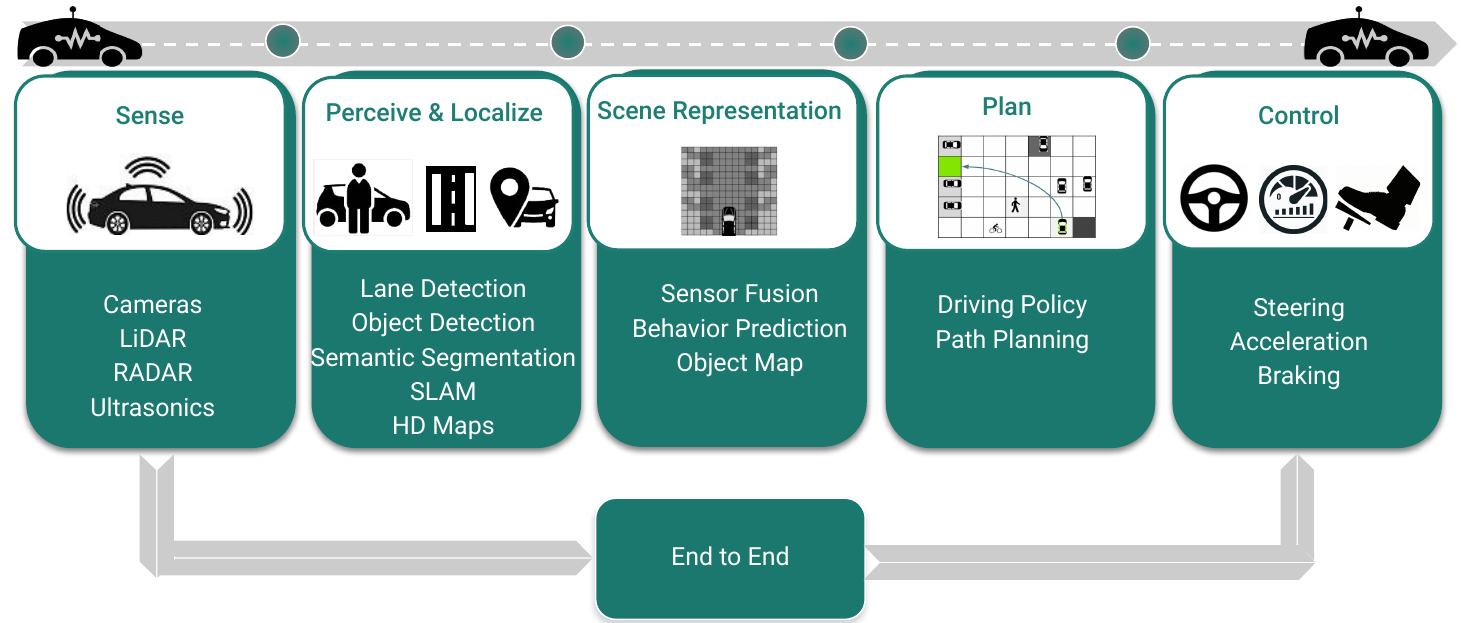}
   \caption{Autonomous Driving Pipeline}
\label{fig:ad_pipeline}
\end{figure*}

When an autonomous vehicle moves from source to destination, a navigator like google maps or HD Maps generate a high-level route. This route is made up of a series of connected nodes at finite distances. The vehicle moves from one node to another in a repetitive way until it reaches the destination. Maneuver occurs in a five-stage process, Sensing, Perception and Localization, Scene Representation, Planning, and Controlling (illustrated in Figure \ref{fig:ad_pipeline}).

In the sensing stage, the vehicle collects information about the surroundings via sensors like Camera, LiDAR, RADAR, and Ultrasonics. Perception involves the extraction of useful information from the raw data like lane positions, presence of pedestrians, and other vehicles   \cite{ sistu2019real, mohapatra2021bevdetnet}, semantic segmentation \cite{chennupati2021learning, chennupati2019auxnet}, moving objects detection   \cite{modnet, mohamed2021monocular}, depth estimation \cite{Kumar2018CVPRWorkshop, kumar2021svdistnet}, feature correspondence \cite{Konrad2021FisheyeSuperPointKD, shen2023optical} and recognition of drivable regions \cite{hughes2019drivespace, stapleton2022neuroevolutionary}. In recent times deep learning algorithms have shown tremendous success in almost all the perception related tasks. Localization is the vehicle's ability to precisely know its position in the real world at decimeter accuracy  \cite{SLAM},   \cite{deep_slam}. In simple words, perception answers what is around the vehicle, and localization answers where is the vehicle precisely. Path planning algorithms  \cite{planning} make use of this related information to define a path to navigate from one node to another. While defining the path, the algorithms use different driving policies like safety, rules of driving, road conditions, and pleasant ride experience for the passengers in the car  \cite{RTT}. These digital instructions from the algorithms are converted into the vehicle's physical movement via the controlling unit  \cite{decision_making}.

\begin{figure}[t]
\centering
    \includegraphics[width=0.95\linewidth]{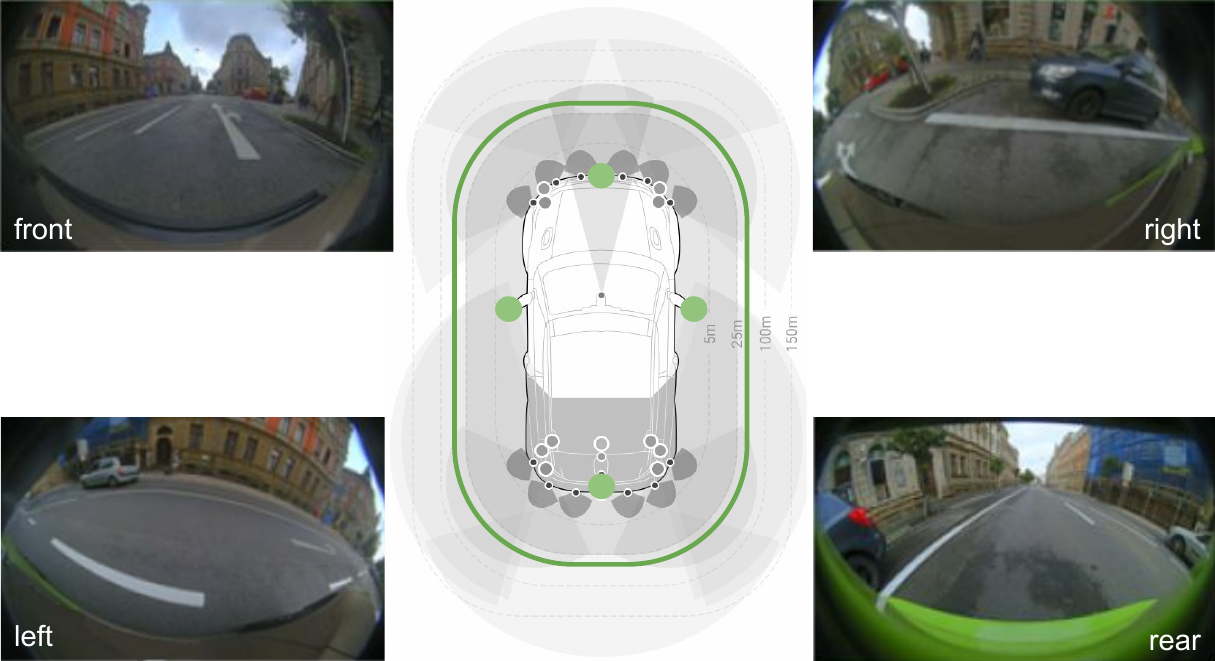}
  \caption{Surround-view camera network images showing near field sensing and wide field of view}
\label{fig:car_new}
\end{figure}
\addtolength{\textfloatsep}{-0.2in}

In the autonomous driving pipeline, perception is a computationally expensive and sophisticated block, and efforts to build unified models for all perception related tasks is an active area of research  \cite{neurall}   \cite{woodscape}. Though there is a considerable debate on what sensors are needed for this task, equivocally, cameras are considered as an essential sensor for any perception pipeline. 
Visual perception is the task of perceiving the information around the vehicle via cameras. A modern automated vehicle consists of anywhere between 4 to 20 cameras of different field of view (FoV) performing different tasks. Visual perception can be described as a combination of Recognition, Reconstruction, and Re-localization. Recognition is knowing what is around the vehicle and involves tasks like Detection, Segmentation, and lens soiling   \cite{soiling, uricar2019let}. Reconstruction consists of depth estimation, motion estimation to know where the objects are in the 3D world. Re-localization knows where the ego vehicle is in the world. It involves pose estimation and SLAM  \cite{SLAM}. Other than this perception also involves lesser-known tasks like trailer angle estimation, measuring sun glare on the lens.

\subsection{Low-Power Hardware for Automated Driving}
Recently demand for low-power SoC based automated vehicles has increased significantly as features like pedestrian detection, emergency braking and lane keep assist started to attract more consumers. Typical low-power SOCs include Renesas V3H, TI TDA4x, and Nvidia Xavier.
The choice of SoC is based on the criteria of performance (Tera Operations Per Second (TOPS), utilization, bandwidth), cost, power consumption, heat dissipation, high to low-end scalability and programmability. The SOC choice provides the computational bounds in the design of algorithms. 
The progress in Convolutional Neural Networks (CNNs) has also led the hardware manufacturers to include their custom hardware computing units to provide a high throughput of over $10$ TOPS targeting Level-3 automation. But Level-2 automation systems still rely on computing units less than $2$ TOPS. In  \cite{yuvmultinet}, authors developed a multitask learning algorithm for hardware with $1$ TOPS of computing power, consuming less than $10$ watts of power. 
\begin{figure*}
\centering
    \includegraphics[width=0.99\linewidth]{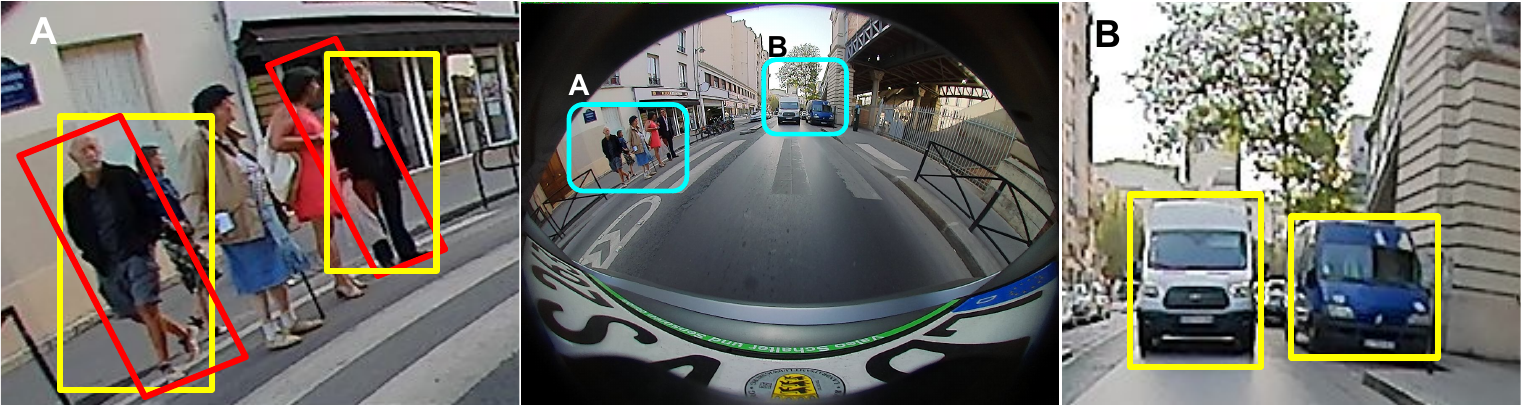}
   \caption{Center: Front camera image. Right(B): Bounding boxes representing objects correctly. Left(A): Bounding boxes and oriented boxes fail to represent objects accurately, more details in Section \ref{Intro}}
\label{fig:bbox_issue}
\end{figure*}
\subsection{Object Detection}
Object detection is the first and foremost problem in visual perception, which involves the recognition and localization of the objects in the image. It has use-cases like emergency braking and collision avoidance etc. Hence object detection models' performance directly influence the success and failure of autonomous driving systems. 
A simple and efficient representation of objects in images is bounding box representation. State-of-the-art methods for object detection based on deep learning can be broadly classified into two types,

\begin{itemize}[nosep]
    \item Two stage detectors
    \item Single-Stage detectors.
\end{itemize}
\subsubsection{Two Stage Object Detection}
In two-stage approaches, object detection is  split into two tasks,
(i) Extraction of Region of Interest (ROI)s and encoding them as features 
(ii) Regressing for bounding boxes in these ROIs using encoded features.
A common practice is to have a high Recall in the first stage to ensure all possible objects like patterns go through the second stage. RCNN\cite{RCNN} was the first to use this approach. In RCNN first stage, a selective search algorithm is used to propose ROIs, followed by a CNN feature extraction. In the second stage, an SVM is trained to classify the objects based on the CNN features.
Unlike RCNN, which extracts CNN features separately on each ROI, Fast-RCNN  \cite{FastRCNN} process the whole image. So the CNN feature extraction is performed only once per image. It has introduced a 25x speed in the inference stage compared to RCNN. Also, Fast-RCNN has replaced SVM with a linear classifier and introduced a linear regressor for bounding box fine-tuning. This moved the Fast-RCNN a step closer to the end differentiable training strategy. Both Fast-RCNN  \cite{FastRCNN} and SPP-net  \cite{he2015spatial} improved RCNN  \cite{RCNN} by extracting RoIs from the feature maps. SPP-net introduced a spatial pyramid pooling (SPP) layer to handle images of arbitrary sizes and aspect ratios. It applies SPP layer over the feature maps generated from convolution layers and outputs fixed-length vectors required for fully connected layers. It eliminates fixed-size input constraints and can be used in any CNN-based classification model. However, Fast-RCNN and SPPnet are not end-to-end trainable as they depend on the region proposal approach. Faster-RCNN  \cite{FasterRCNN} solved this limitation by introducing Region Proposal Network (RPN), which made end-to-end training possible. RPN generates RoIs by regressing a set of reference boxes, known as anchor boxes. This introduced two streams for object detection, i.e., a common encoder and two decoders. The efficiency of Faster-RCNN is further improved by RFCN  \cite{RFCN}, which replaces fully connected layers with fully convolutional layers.

\subsubsection{Single Stage Object Detection}

These approaches eliminate the RoI extraction stage and carry out classification and regression of bounding boxes directly on CNN feature maps and hence a single state encoder-decoder style network performs localization and classification tasks. Overfeat \cite{overfeat} proposed a unified framework to perform two tasks: classification and localization using a multi-scale, sliding window approach. YOLO  \cite{YOLOV1} divides the input image into grids and predicts bounding boxes directly by regression and classification at each grid. This soon became a defacto style for single state real-time object detection on low power hardware like mobile phones and Level-3 autonomous driving engines. YOLO9000 (YOLOv2)  \cite{YOLO9000} improved the performance by introducing batch normalization and replacing fully connected layers of YOLOv1 with anchor boxes for bounding box prediction. Anchor boxes are computed over the dataset, representing the average variation of height and width of the objects in the dataset. Instead of directly regressing for object width and height, YOLOv2 predicts off-sets from a predetermined set of anchors with particular height-width ratios. YOLOv3  \cite{YOLOV3}, a faster and accurate object detector than previous versions, uses Darknet-53 as its feature extraction backbone. YOLOv3 can detect small objects with a multi-scale prediction approach, a significant drawback in earlier versions. \\
Single Shot Multibox Detection(SSD)  \cite{ssd} places dense anchor boxes over the input image and extract feature maps from multiple scales. It then classifies and regresses the bounding boxes relative to anchor boxes.
DSSD  \cite{dssd} replaced the VGG network of SSD with Residual-101. It is then augmented with a deconvolution module to integrate feature maps from the early stage with the deconvolution layers. It outperforms the SSD in detecting small objects. MDSSD  \cite{mdssd} further extends DSSD with fusion blocks to handle feature maps at different scales. 
\\
RetinaNet  \cite{RetinaNet} introduced focal loss to address foreground and background class imbalance during training. It matches or surpasses the accuracy of state-of-the-art two-stage detectors while running at faster speeds. The architecture shares ‘anchors’ from RPN and builds a single Fully Convolutional Network (FCN) with Feature Pyramid Network (FPN) on top of the ResNet backbone.
\begin{figure*}
\centering
    \includegraphics[width=0.99\linewidth]{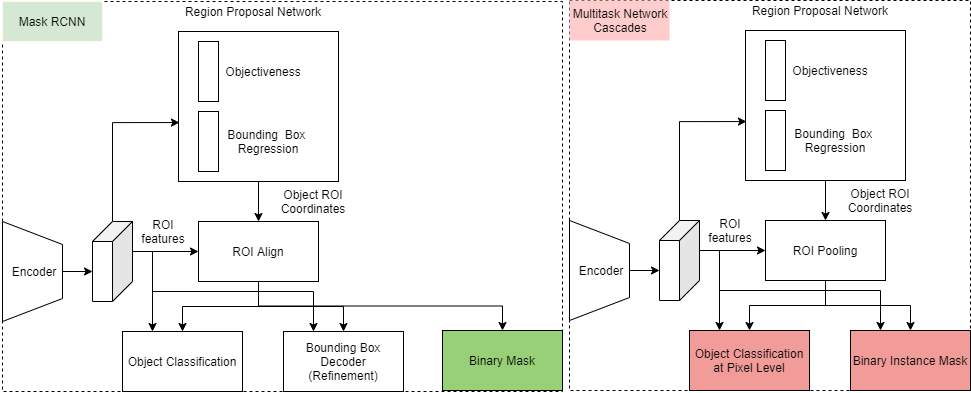}
   \caption{Comparison between MaskRCNN and Multi-task Network Cascade. Both models are two stage approaches and use FasterRCNN components (blocks not colored)}
\label{fig:mask_vs_faster}
\end{figure*}

\subsection{Instance Segmentation}
Instance segmentation involves predicting both object bounding box and pixel-level object mask.

\subsubsection{Two stage Instance Segmentation}
Intuitively instance segmentation can be modeled as a bounding box detection followed by a binary segmentation within the box. This paradigm is referred to as 'Detection then Segmentation'. Models following this approach often achieve a state of the art performance but are quite slow to adapt to real-time applications. MaskRCNN  \cite{MaskRCNN} adapted this approach by using FasterRCNN for bounding box detection and an additional decoder for object mask segmentation. Here segmentation is performed as a binary classification to differentiate object pixels from the background or other object pixels. Multi-task Network Cascade (MNC)  \cite{MNC} uses a similar approach to MaskRCNN. It uses RPN for box proposals, followed by class agnostic instances generation on the proposed regions and finally categorical classification of these mask instances. Figure \ref{fig:mask_vs_faster} shows the similarity between MNC and MaskRCNN algorithms. During the inference time on a 12 GB, 7 TFLOPs NVIDIA M40 GPU, MaskRCNN reported a ~6 FPS run time. Today even the Level-5 autonomous vehicles use only 1.3 to 2 TFLOPs computing engines for running the complete deep learning stack, making a state of the art two-stage approach far from reality for L3 automated vehicles. This led to a recent trend of simplistic single-stage object detection style instance segmentation techniques like PolarMask  \cite{polarmask} YOLOACT  \cite{yoloact} and PolyYOLO  \cite{polyyolo}.
\subsubsection{Single stage Instance Segmentation}

YOLOACT  \cite{yoloact} uses a single encoder dual parallel decoder style architecture for instance-level image segmentation. Encoder is same as RetinaNet backbone, i.e Feature Pyramid Network with Resnet101  \cite{he2016deep}. The first decoder generates a set of \textit{k} prototype masks at image resolution. These masks do not depend on any single object class. However, these masks represent instance masks of an object when multiplied with the correct set of coefficients. The second decoder is a standard bounding box decoder with extra computation to predict mask coefficients for each object instance. Instance masks for objects are generated as a linear combination of prototype masks and mask coefficients. Though YOLOACT performance is lower than MaskRCNN, it is 5x faster in run time.
\subsubsection{Polygon Instance Segmentation}
 PolarMask \cite{polarmask} and PolyYOLO  \cite{polyyolo} regress for contour boundaries in polar space. It is hence removing computational overheads of an extra decoder and segmentation of pixels at images level. 

Other approaches to instance segmentation range from clustering of instance embedding   \cite{proposal_free},   \cite{joint_instance} to prediction of instance centers using offset regression   \cite{pan_deplab}. These methods appear intuitively designed but are lagging in terms of accuracy and computational efficiency. The major drawback of these methods is the usage of compute-intensive clustering methods like OPTICS   \cite{OPTICS}, DBSCAN  \cite{density_cluster}.


\section{Object Detection on Fisheye Cameras}
\begin{figure}[!t]
   \centering
   \includegraphics[width=0.99\linewidth]{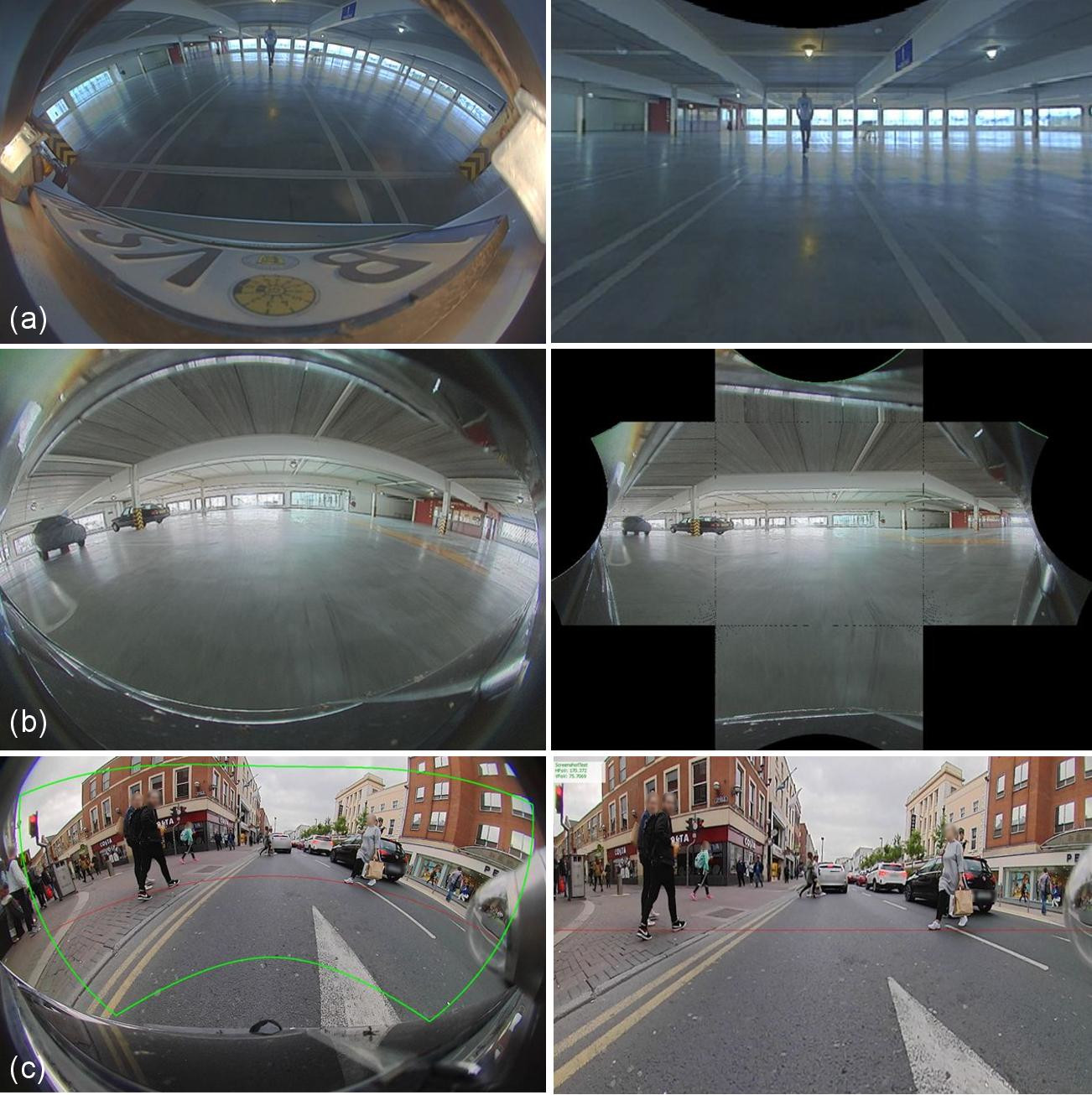}
    \caption{Undistorting the fisheye image:
    (a) Rectilinear correction; (b) Piecewise linear correction; (c) Cylindrical correction. Left: raw image; Right: undistorted image.}
\label{fig:projections}
\end{figure}
Fisheye cameras make use of non-linear mapping to generate a large field of view. With just four surround-view fisheye cameras, we can achieve a dense $360\degree$ near field perception, making them suitable for automated parking, low-speed maneuvering, and emergency braking. A commercial fisheye camera usually has a $190\degree$ horizontal field of view as shown in Figure \ref{fig:car_new}. It is usually available from 2MP to 20MP resolution. However, this advantage comes at the cost of non-linear distortions in the image. Objects at different angles from the project center look quite different, making the object detection a challenge.

A common practice is to rectify distortions in the image by a 4\textsuperscript{th} order polynomial model or Unified camera model \cite{kumar2023surround, eising2021near}. The fact is, there is no ideal projection or correction. These corrections are application-driven, and every correction technique has its disadvantages (Figure \ref{fig:projections}). Rectilinear correction suffers from loss of Field of View (FoV) and sampling issues, Piece-wise linear with artifacts at transition areas and massive bleeding in the image, and Cylindrical as a quasi-linear correction, offers a practical trade-off.  Another overhead is extra computational resources needed for correction as the visual perception pipeline usually have different algorithms demanding different view projections. Though Look Up Tables (LUTs) make this correction process accelerated, LUTs rely on online calibration that needs to be generated every time there is a change in the online calibration.

Despite these disadvantages, image correction is encouraged due to the limitations of the non or early deep learning object recognition and segmentation algorithms. With a push in deep neural networks, this trend is slowly changing. Modern CNN based object detection algorithms like YOLO and FasterRCNN can detect objects on raw fisheye images and main issue with object detection on raw fisheye images is representation of objects as bounding boxes.\par
\subsection{Bounding boxes on Fisheye}

Objects go though serious deformations due to radial distortion in fisheye images and box representation fails in many practical scenarios \cite{rashed2020fisheyeyolo}. Here are two scenarios where the correct representation of objects is as important as the detection.\par
\subsubsection{Pedestrian Localization Issue}
In Figure \ref{fig:bbox_issue}.B, vehicles are near the center region of the image, and hence the lower part of the bounding boxes represent the object intersection with the road quite well. However, in Figure \ref{fig:bbox_issue}.A, standard bounding boxes in yellow color are not good enough to represent the object road intersection.\par
The common idea is to orient the boxes as shown in red color in Figure \ref{fig:bbox_issue}.A. In the case of the person on the left side, this orientation concept works. As the box with optimal orientation is also a box with optimal IoU with the ground truth. However, in the case of the person in a black suit, the optimally oriented box is not the optimal IoU. So simple orientation works in some cases, but it does not solve the problem. 3D boxes work, but both annotating and inferring a 3D box is a noisy process for small objects.\par
\subsubsection{Missing Parking Spot}
\begin{figure*}
\centering
    \includegraphics[width=0.99\linewidth]{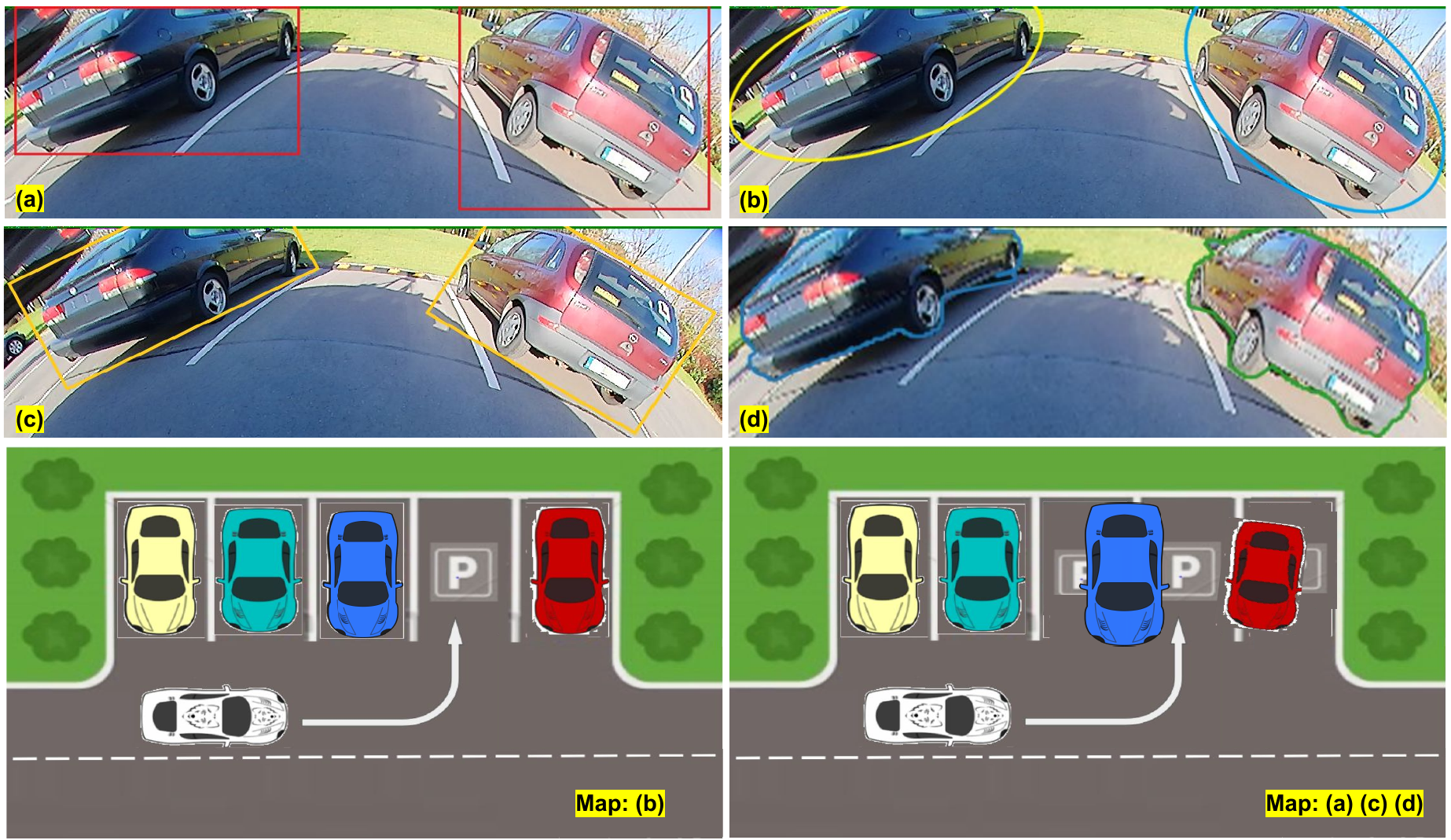}
  \caption{Parking spot failure case. Bottom left: 2D map of navigation showing free parking spot. Bottom right: Same 2d map showing no free parking spot. Object detection as bounding box(a), ellipse(b), oriented box (c) and instance segmentation (d)}
\label{fig:parking_failure}
\end{figure*}

A correctly detected but improperly represented objects can result in failure cases like missing a parking spot or in non-optimal path planning. Figure \ref{fig:parking_failure} shows an automated car maneuvering to a parking slot between the two cars. Two cars got detected by bounding box, oriented box, ellipse and polygon object detection algorithms. However, only in instance segmentation case objects are located correctly outside the free parking spot. In rest of the cases, objects seems to be present inside the parking spot and in those cases the free parking spot in maneuver mapping shows as occupied (bottom row images). This shows that the detection of objects is as important as correct representation in fisheye based visual navigation systems.
\par
A full-fledged solution to this problem is instance segmentation, but most state-of-the-art algorithms like MaskRCNN demand higher computing powers and are unrealistic to work on low-power hardware that is generally used in Level2 and Level 3 autonomous vehicles. Hence there is a need for memory and computation efficient models. 
It encouraged us to develop FisheyeDetNet, a single network to perform object detection and instance segmentation to deploy on low power hardware accelerators. It is an efficient, small footprint network that uses ResNet18 as a backbone and YOLO style head for polygon based instance segmentation.
\begin{table*}[t]
\centering
\begin{tabular}{|l|l|l|l|l|l|l|l|}
\hline
\rowcolor[HTML]{EFEFEF} 
{\color[HTML]{000000} Representation} & {\color[HTML]{000000} Bounding Box} & {\color[HTML]{000000} Rotated Box} & {\color[HTML]{000000} 12 points} & {\color[HTML]{000000} 24 points} & {\color[HTML]{000000} 36 points} & {\color[HTML]{000000} 60 points} & {\color[HTML]{000000} 120 points} \\ \hline
\rowcolor[HTML]{FFFFFF} 
IoU & \multicolumn{1}{r|}{\cellcolor[HTML]{FFFFFF}0.552} & \multicolumn{1}{r|}{\cellcolor[HTML]{FFFFFF}0.643} & \multicolumn{1}{r|}{\cellcolor[HTML]{FFFFFF}0.853} & \multicolumn{1}{r|}{\cellcolor[HTML]{FFFFFF}0.897} & \multicolumn{1}{r|}{\cellcolor[HTML]{FFFFFF}0.918} & \multicolumn{1}{r|}{\cellcolor[HTML]{FFFFFF}0.942} & \multicolumn{1}{r|}{\cellcolor[HTML]{FFFFFF}0.984} \\ \hline
\end{tabular}
\caption{Upper bound on performance of various representations}
\label{table:rep_vs_iou}
\end{table*}

\section{Proposed Method}
\begin{figure*}
\centering
    \includegraphics[width=0.90\linewidth]{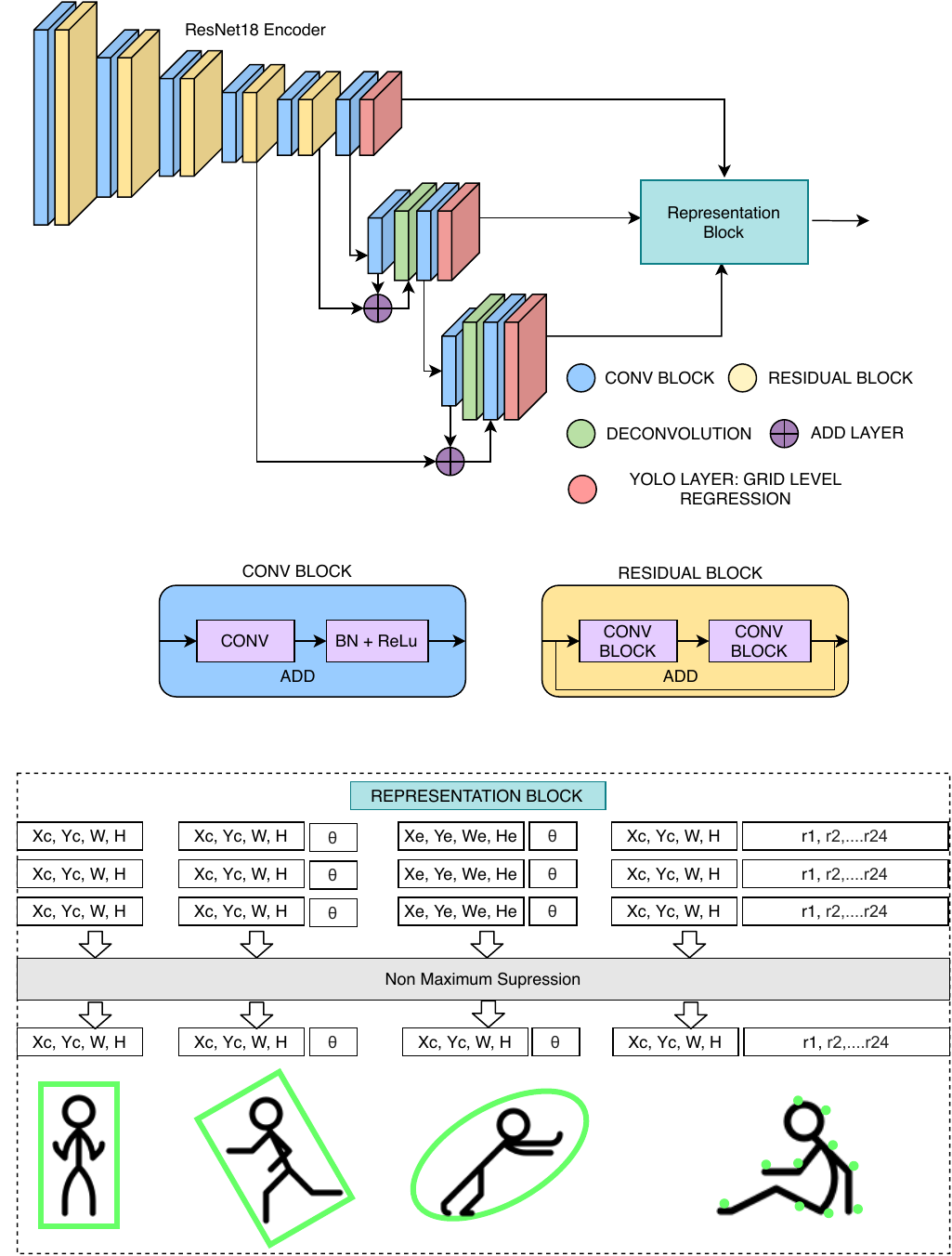}
   \caption{Proposed object detection network architecture and comparison between different representations. Total four models tested and benchmarked on Valeo fisheye dataset}
\label{fig:proposed_network}
\end{figure*}

Objects detection can be represented as bounding boxes, rotated bounding boxes, ellipses, and polygons. Irrespective of the algorithm used, each representation has a limitation on maximum performance it can achieved on the given dataset. We term this as empiricall upper bound' or simply upper bound. The same is shown in Table \ref{table:rep_vs_iou}, where mean IoU score between the annotations from each representation and ground truth instance annotations is presented for our fihseye dataset. In case of polygon representation, instance annotations are generated by sampling 12, 24, 36, 60 and 120 points per $360\degree$ in polar coordinates. 
We modified the YOLOv3  \cite{YOLOV3} network to accommodate all these four different representations.
\begin{itemize}[nosep]
    \item Bounding Box
    \item Oriented Bounding Box
    \item Ellipse 
    \item Polygon
\end{itemize}
To make the network feasible to port onto a low power automotive hardware, we used ResNet18   \cite{he2016deep} as an encoder. Compared to standard Darknet53 encoder  \cite{YOLOV3}, this has nearly 60\% fewer parameters. Proposed network architecture is shown in Figure \ref{fig:proposed_network}. Different representations are implemented in representation block.

\subsection{Bounding Box}
Our Bounding box model is the same as YOLOv3 except Darknet53 encoder is replaced with ResNet18 encoder. Similar to YOLOv3, object detection is performed at multiple scales. For each grid in each scale, object width($\hat{w}$), height($\hat{h}$), object center coordinates($\hat{x}$, $\hat{y}$) and object class is predicted. Finally, a non-maximum suppression is used to filter out the redundant detections. Instead of using $L_{2}$ loss for categorical and objectness classification, we used standard categorical cross-entropy and binary entropy losses, respectively.\par
Representing the modified YOLO loss as a combination of sub-losses, 
\begin{align}
\mathcal{L}_{xy} &= \lambda_{coord} \sum_{i=0}^{S^2}\sum_{j=0}^{B} l_{ij}^{obj}[(x_{i}-\hat{x}_{_{i}})^2 + (y_{i}-\hat{y}_{_{i}})^2]\\
\mathcal{L}_{wh} &= \lambda_{coord} \sum_{i=0}^{S^2}\sum_{j=0}^{B} l_{ij}^{obj}[(\sqrt{w_{i}}- \sqrt{\hat{w}_{_{i}}})^2 \nonumber \\   
&\quad \quad \quad \quad \quad  \quad\quad\quad  + (\sqrt{h_{i}}-\sqrt{\hat{h}_{_{i}}})^2]  
\end{align}

\begin{align}
\mathcal{L}_{obj} &= -\sum_{i=0}^{S^2}\sum_{j=0}^{B} C_{i}log(\hat{C}_{_{i}})  \\ 
\mathcal{L}_{class} &=  -\sum_{i=0}^{S^2} l_{ij}^{obj}\sum_{c=classes} c_{i,j}log(p(\hat{c_{_{i,j}}})) \\
\mathcal{L}_{total} &= \mathcal{L}_{xy} + \mathcal{L}_{wh} + \mathcal{L}_{obj} + \mathcal{L}_{class}
\end{align}
where height and width are predicted as offsets from pre-computed anchor boxes.
\begin{align}
    \hat{w} &= a_{w} * e^{fw} \\
    \hat{h} &= a_{h} * e^{fh} \\
    \hat{x} &= g_{x} + f_{x} \\
    \hat{h} &= g_{y} * f_{y}
\end{align}
where $a_{w}$, $a_{h}$ anchor box width and height. $f_{w}$, $f_{h}$, $f_{x}$, $f_{y}$ are the outputs from last layer of the network at grid location $g_{x}$, $g_{y}$.

\subsection{Oriented Bounding Box}
In this model along with the regular box information ($\hat{w}$, $\hat{h}$, $\hat{x}$, $\hat{y}$), orientation of the box $\hat{\theta}$ is also regressed. Orientation ground truth range (-180 to +180\degree) is normalized between -1 to +1. The loss function is same as the regular box loss but with an additional term for orientation loss. 
\begin{align}
\mathcal{L}_{orn} &= \sum_{i=0}^{S^2}\sum_{j=0}^{B} l_{ij}^{obj}[\theta_{i}-\hat{\theta}_{_{i}}]^2 \\
\mathcal{L}_{total} &= \mathcal{L}_{xy} + \mathcal{L}_{wh} + \mathcal{L}_{obj} + \mathcal{L}_{class} + \mathbf{\mathcal{L}_{orn}}
\end{align}
where $\mathcal{L}_{total}$, is the total loss minimized for oriented box regression. 
\subsection{Ellipse Detection}
Ellipse regression is the same as oriented box regression. The only difference is in the output representation. Hence the loss function is also the same as oriented boxes loss.
\begin{figure}[t]
\centering
    \includegraphics[width=0.95\linewidth]{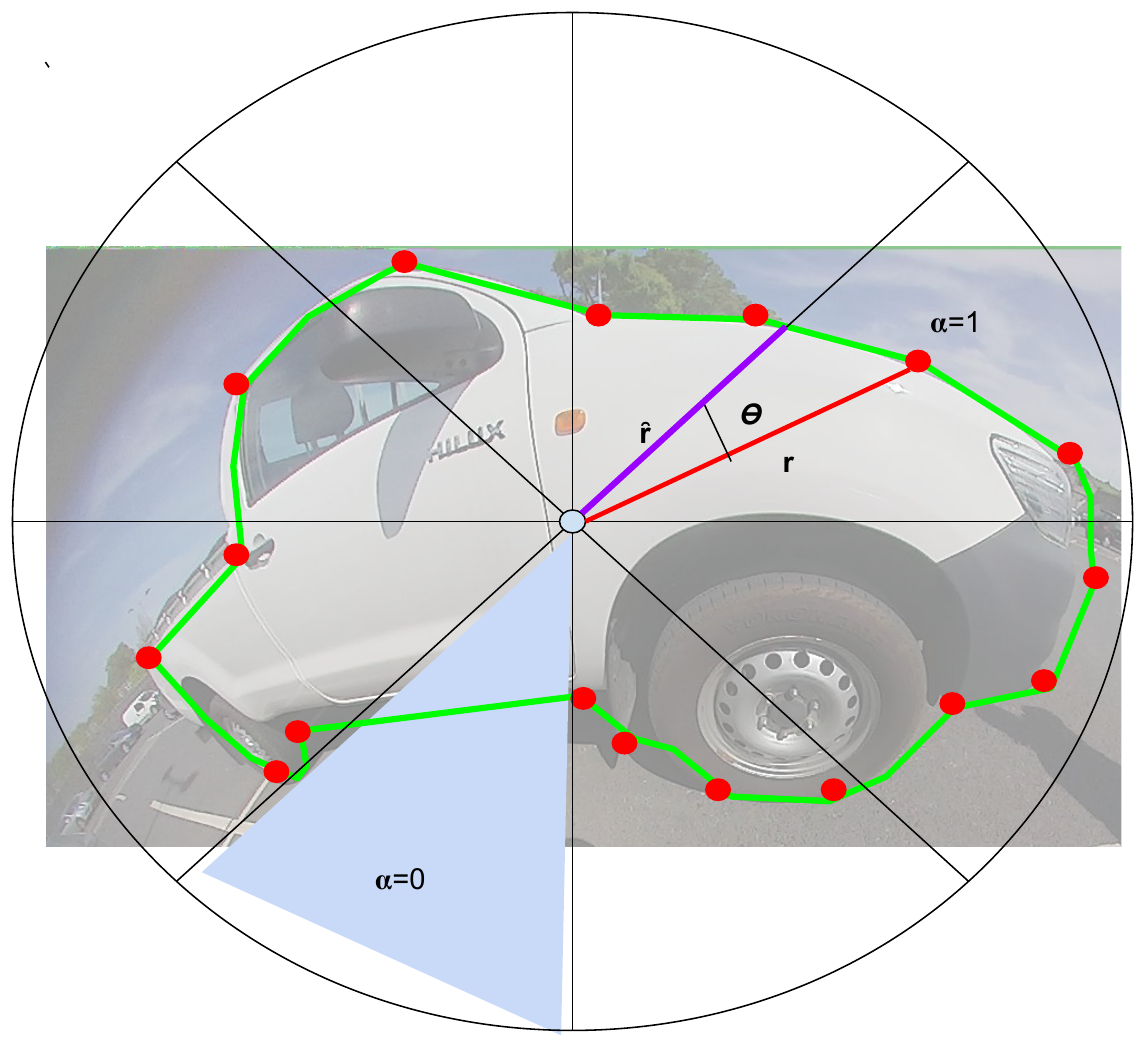}
   \caption{Dense pixel level annotation sampling (purple) vs sparse polygon points annotation sampling (red) in polar space.}
\label{fig:sparse_vs_dense}
\end{figure}

\begin{table*}[t]
\centering
\begin{tabular}{
>{\columncolor[HTML]{EFEFEF}}c 
>{\columncolor[HTML]{FFFFFF}}c 
>{\columncolor[HTML]{FFFFFF}}c 
>{\columncolor[HTML]{D9D9D9}}c 
>{\columncolor[HTML]{FFFFFF}}c 
>{\columncolor[HTML]{FFFFFF}}c 
>{\columncolor[HTML]{D9D9D9}}c }
\hline
\multicolumn{1}{|c|}{\cellcolor[HTML]{EFEFEF}} & \multicolumn{3}{c|}{\cellcolor[HTML]{C0C0C0}Representation vs Representation} & \multicolumn{3}{c|}{\cellcolor[HTML]{C0C0C0}Representation vs Instance Annotation} \\ \cline{2-7} 
\multicolumn{1}{|c|}{\multirow{-2}{*}{\cellcolor[HTML]{EFEFEF}Experiment}} & \cellcolor[HTML]{EFEFEF}Vehicle & \cellcolor[HTML]{EFEFEF}Pedestrian & \cellcolor[HTML]{EFEFEF}mAP & \cellcolor[HTML]{EFEFEF}Vehicle & \cellcolor[HTML]{EFEFEF}Pedestrian & \cellcolor[HTML]{EFEFEF}mAP \\ \hline
Bounding Box & 0.6627 & 0.3157 & 0.4892 & 0.5132 & 0.3150 & 0.4141 \\ \hline
Oriented Box & 0.6548 & 0.3010 & 0.4779 & 0.5234 & 0.3185 & 0.4210 \\ \hline
Ellipse & 0.6601 & 0.2900 & 0.4751 & 0.5290 & 0.2889 & 0.4090 \\ \hline
Polygon (24 points) & 0.6624 & 0.3140 & 0.4882 & 0.6761 & 0.3155 & 0.4958 \\ \hline
\end{tabular}
\caption{Comparison of various representations. In case of polygon experiment, Representation vs Representation metric is between a bounding box annotation and bounding box predicted along with polygons}
\label{table: results}
\end{table*}
\subsection{Polygon Detection}
Our proposed approach for polygon-based instance segmentation is quite similar to PolarMask  \cite{polarmask} and PolyYOLO   \cite{polyyolo} approaches. Instead of using sparse polygon points and single scale predictions like PolyYOLO. We use dense polygon annotations and multi-scale predictions. Instead of heavy backbone architecture like PolarMask, we employed lightweight ResNet-18 as our encoder. These changes enabled us to develop a small footprint instance segmentation model with just ~13M parameters. As there is no heavy encoder backbone or feature map upscaling to image level and segmentation at the pixel level, our model is quite suitable for real-time applications like object detection on Level-3 automotive ECUs. Keeping the network architecture similar in all the four experiments results in a fair comparison between different representations. 

\begin{figure*}
\centering
    \includegraphics[width=0.95\linewidth]{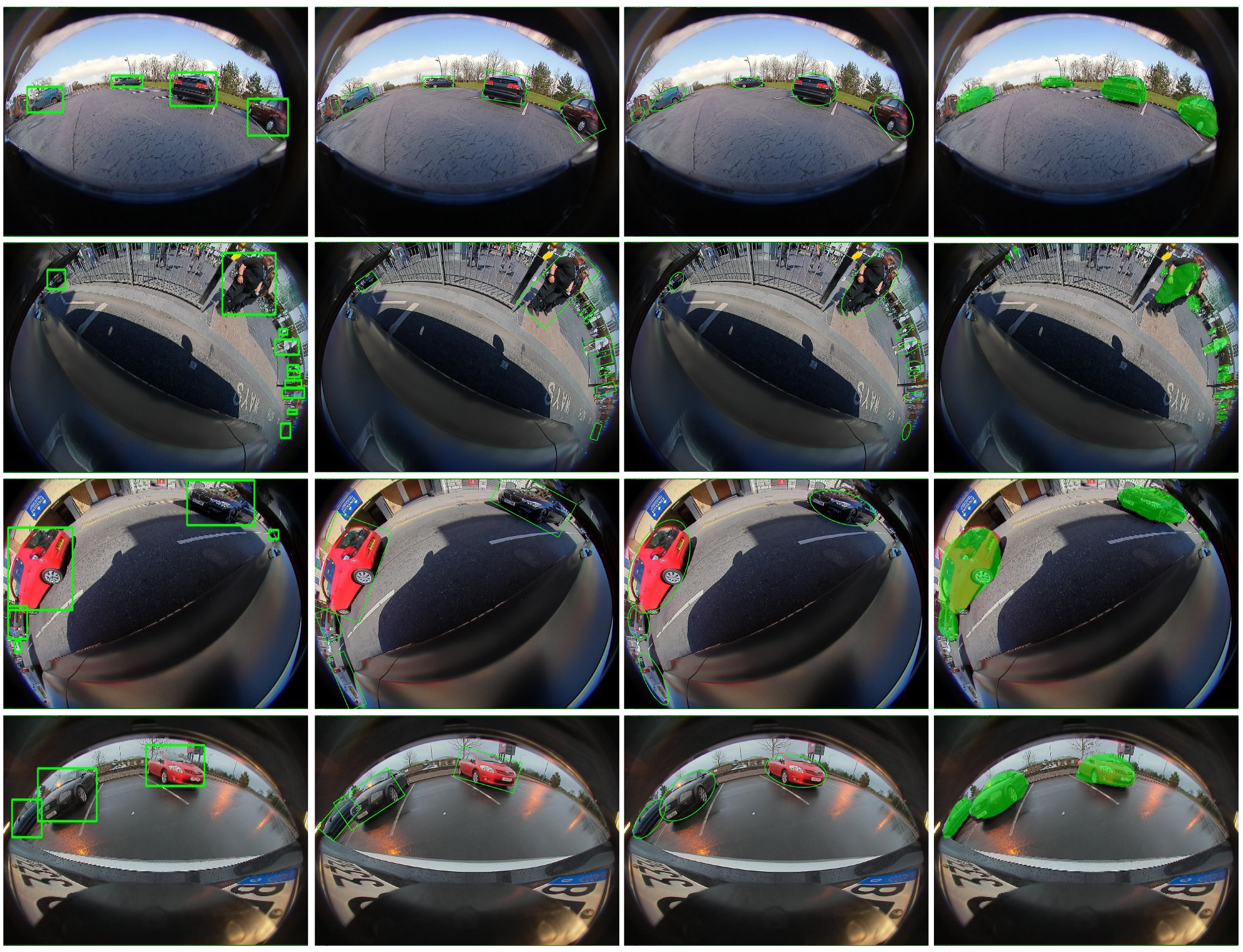}
   \caption{Qualitative results: Each row shows output of all four models on the same image. Each column shows images captured by FV, MRV, MLV and RV cameras on the vehicle}
\label{fig:qual_results}
\end{figure*}

The polygon regression loss is given by,
\begin{align}
\mathcal{L}_{poly} &= \sum_{i=0}^{S^2}\sum_{j=0}^{B} \sum_{k=0}^{R} l_{ij}^{obj}[r_{i,k}-\hat{r}_{_{i,k}}]^2 \\
\mathcal{L}_{total} &= \mathcal{L}_{xy} + \mathcal{L}_{wh} + \mathcal{L}_{obj} + \mathcal{L}_{class} + \mathbf{\mathcal{L}_{poly}}
\end{align}
The total loss is given by $\mathcal{L}_{total}$, where R corresponds to the number of sampling points, each point is sampled with a step size of $360/R$ angle in polar coordinates, as shown in Figure \ref{fig:sparse_vs_dense}. We used dense pixel-level annotations, and hence there is only one parameter needed to represent each polygon point in the polar coordinate system. It is similar to PolarMask. PolyYOLO, on the other hand, uses sparse polygon points (in red), and thus requires 3 parameters $r$, $\theta$ and $\alpha$. Hence the total required parameters for R sampling points are 3*R in case of sparse polygon points-based annotations. 
The effect of different sampling rates w.r.t actual pixel-level annotation masks is presented in Table \ref{table:rep_vs_iou}.


\section{Experiments}
All the four models are systematically tested on large scale automotive fisheye surround-view dataset. Dataset deatils, metrics and evaluation criteria and training details are presented in the following subsections. 

\subsection{Dataset}
\begin{figure*}
\centering
    \includegraphics[width=0.95\linewidth]{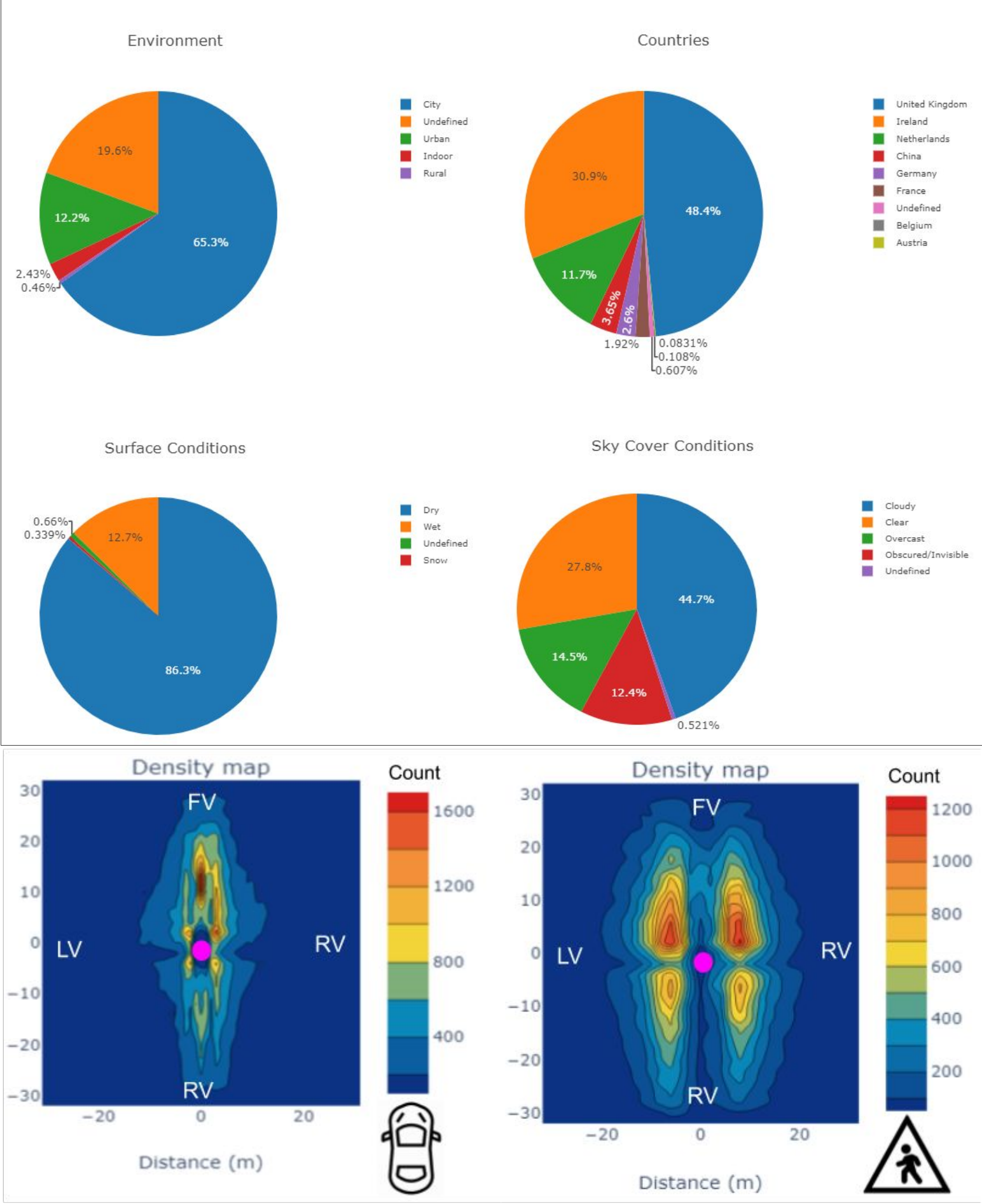}
   \caption{Dataset Statistics: Pink Dot in density maps is an ego vehicle center. FV: Front View, RV: Rear View, MLV: Mirror Left View, and MRV: Mirror Right View w.r.t the ego vehicle. The pie charts show the diversity in geographical and climatic conditions.}
\label{fig:dataset_stats}
\end{figure*}

We presented results on Valeo proprietary dataset for automated driving applications. This dataset comprises of 50,000 images captured from four surround-view cameras \cite{uricar2019challenges, yahiaoui2019overview}. Instance segmentation for vehicles and pedestrians. A subset of this dataset with more annotation classes and annotation for different tasks is presented in   \cite{woodscape, ramachandran2021woodscape}. Figure  \ref{fig:dataset_stats} shows the diversity of geographical, climatic conditions of the dataset. The density maps show that majority of the vehicles and pedestrians are within 20 meters of the vehicle. It is an important metric as fisheye surround-view cameras are usually mounted for near filed visual perception applications. Dataset is divided into train, validation, and test splits at 70, 15, 15 proportions. A random sampling technique is used for this purpose.

\subsection{Training Details}
All four models are trained on nearly 35K images at an input resolution of 544X288 $(width X height)$. A pre-trained ResNet18 model without classification layers is used as Encoder and horizontal image flip as data augmentation technique. All models are developed on PyTorch v1.4 \cite{pytorch}. Training, evaluation and inference are performed on a NVIDIA GTX 1080Ti GPU. All models are trained for 80 epochs with early stopping criteria based on validation loss. Ranger optimizer \cite{ranger} and one cycle learning rate scheduler  \cite{one_cycle} is used for optimization. Ranger uses gradient stabilization, combines RAdam \cite{r_adam} and LookAhead \cite{lookahead} in one optimizer. Hence helps in stabilized training. 

\subsection{Results}
All models are compared using a mean average precision metric (mAP) with an IoU threshold of $50\%$. Results are presented in Table \ref{table: results}. Each algorithm is evaluated based on two criteria - Performance on same representation and on instance segmentation. For example, a bounding box detection model predictions are compared with bounding box ground truth (Representation vs Representation) and instance mask ground truth (Representation vs Instance Annotation). While the comparison with the same representation shows the performance of the algorithm, comparison with instance masks shows its closeness to its upper bound. Results in Table \ref{table: results} are in alignment with the empirical upper bounds shown in Table \ref{table:rep_vs_iou}. This shows that many practical failure cases like missing parking spots can be solved with a change in representation as opposite to increasing network capacity. 
\\
Qualitative results on test set for all four representations on all four cameras are shown in Figure \ref{fig:qual_results}, In Row-1: Though all four models detected the vehicles, polygon segmentation is the only representation to solve the missing parking spot problem. In Row-2: Oriented boxes and ellipse are able to locate the pedestrian precisely, while standard box and polygon failed. Row-4: Missing parking spot problem is handled well by both ellipse and polygon segmentation representation models. 
\section{Conclusion}
In this work, we studied the problem of bounding box object detection on fisheye images. First, we demonstrated that due to strong radial distortions the bounding box is not a good representation of object detection on fisheye images due to strong radial distortion. Then, we explored several improved representations starting from a rotated bounding box, ellipse, and then finally a generic polygon. We proposed a novel algorithm by extending YOLO to regress a generic representation across the representations, as mentioned above. We call our algorithm FisheyeDetNet, and the implementation demonstrates significant improvements over the baseline representations. We also showed that many of the practical problems can be solved by learning the right representations instead of increasing the model complexity with same models. 

\bibliographystyle{IEEEtran}
\bibliography{egbib}

\end{document}